# Gallbladder Cancer Detection in Ultrasound Images based on YOLO and Faster R-CNN

Sara Dadjouy
*Department of Mathematics, Statistics and Computer Science*
*College of Science, University of Tehran*
Tehran, Iran
sara.dadjouy@ut.ac.ir

Hedieh Sajedi
*Department of Mathematics, Statistics and Computer Science*
*College of Science, University of Tehran*
Tehran, Iran
hhsajedi@ut.ac.ir

*Abstract*— Medical image analysis is a significant application of artificial intelligence for disease diagnosis. A crucial step in this process is the identification of regions of interest within the images. This task can be automated using object detection algorithms. YOLO and Faster R-CNN are renowned for such algorithms, each with its own strengths and weaknesses. This study aims to explore the advantages of both techniques to select more accurate bounding boxes for gallbladder detection from ultrasound images, thereby enhancing gallbladder cancer classification. A fusion method that leverages the benefits of both techniques is presented in this study. The proposed method demonstrated superior classification performance, with an accuracy of 92.62%, compared to the individual use of Faster R-CNN and YOLOv8, which yielded accuracies of 90.16% and 82.79%, respectively.

*Keywords—Artificial Intelligence ،Object Detection ،Ultrasound ،Gallbladder Cancer ،YOLO ،Faster R-CNN*

## I. Introduction

The use of Artificial Intelligence (AI) in disease diagnosis, especially from medical image analysis, is rapidly expanding and advancing. Artificial intelligence algorithms have the capacity to extract and process vast amounts of data quickly and accurately. This capability not only reduces human error but also increases work efficiency, lightens the workload of healthcare staff, and cuts costs. Most importantly, it ensures improved treatment and care for patients [1].

In the realm of medical image analysis for disease diagnosis, one of the main steps is the detection of the Region of Interest (ROI). By focusing on specific regions, computational requirements can be reduced and distractions from irrelevant information will be minimized, resulting in enhanced efficiency and accuracy [2].

Regions of Interest (ROIs) in images can be automatically identified using object detection algorithms. They primarily consist of two steps: feature extraction, and the classification and location identification of objects. Object detection algorithms are broadly divided into two categories: two-stage and one-stage. Two-stage algorithms first generate a set of region proposals, which are areas in the image that could potentially contain an object. These regions are then classified into different object categories and their locations are refined. Conversely, one-stage algorithms simplify the process by eliminating the need for region proposals. They directly predict the class and location of objects in a single step. Popular examples of these algorithms include Faster R-CNN for two-stage detection and YOLO for one-stage detection [3] [4].

The application of these AI techniques is particularly crucial in diagnosing diseases like gallbladder cancer, an aggressive and lethal disease for which early diagnosis is vital [5]. Ultrasound imaging, chosen for its safety, cost-effectiveness, and ease of use, is a popular and primary modality used in the diagnosis of this cancer. However, diagnosing gallbladder diseases can be challenging due to complex conditions that can impact the effectiveness of ultrasound [6]. In the absence of suspicion, cancer can progress undetected as additional tests are typically not conducted. This underscores the importance of enhancing our understanding and interpretation of ultrasound images for gallbladder cancer detection [7].

In this study, the effectiveness of Faster R-CNN and YOLOv8 in detecting gallbladder from ultrasound images and their impact on gallbladder cancer classification are evaluated. Additionally, a fusion method of these techniques is presented, aimed at enhancing the detection and classification accuracy.

## II. Overview of Object Detection Techniques

### A. Faster R-CNN

The Region-based Convolutional Neural Network (R-CNN) is the first model for two-stage object detection that utilizes Convolutional Neural Network (CNN). It identifies region proposals of an image using a selective search algorithm and then applies a CNN model to each of them for classification. It has several limitations, including a stepwise training procedure, a significant demand for both temporal and spatial resources, and slow processing speed. Fast R-CNN overcomes these limitations. It is designed to be trained in a single-step procedure. The algorithm instead of processing each region proposal individually, processes the entire image at once, enhancing its speed and efficiency.

Unlike its predecessors, Faster R-CNN does not depend on an external region proposal method, which can be time-consuming. Instead, it employs a Region Proposal Network (RPN) that efficiently generates candidate region proposals within the image. Each proposal is assigned an 'objectness' score that indicates the likelihood of it containing an object. These scores are used to select relevant region proposals by applying a threshold [4]. Fig. 1 shows the structure of Faster R-CNN.

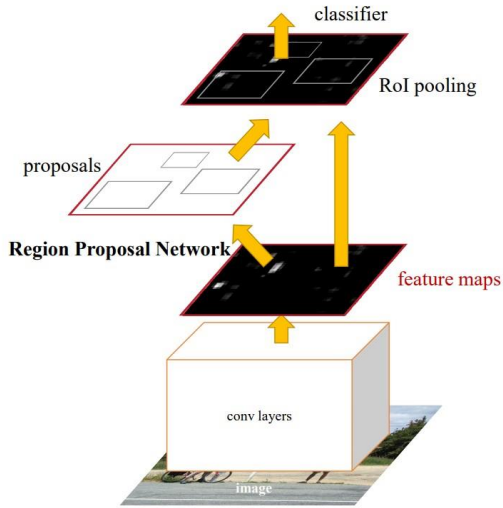

Fig. 1. The structure of Faster R-CNN [8].

## B. YOLO

The You Only Look Once (YOLO) model is a one-stage object detector that employs a single CNN to directly detect bounding boxes and classify them in an input image. The image is partitioned into a predetermined number of grids, with each grid cell predicting a certain number of bounding boxes, along with a confidence score. This score reflects both the model's certainty about the presence of an object within the corresponding bounding box and how well it fits compared to the ground truth. Bounding boxes with class probabilities that exceed a certain threshold are selected for further object localization [4]. Fig. 2 shows an example of how YOLO operates on an image.

Since its introduction, multiple versions of YOLO have been released, each offering its own improvements and enhancements. The latest version, YOLOv8, was released by Ultralytics in January 2023 and will be utilized in the experiments conducted for this paper.

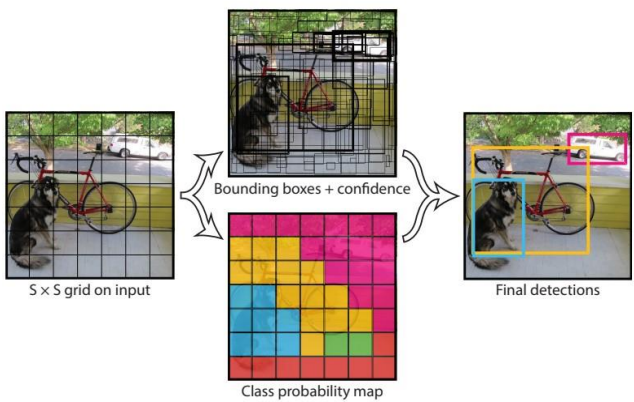

Fig. 2. Illustration of the YOLO procedure applied to images [9].

## C. Comparison of Faster R-CNN and YOLO

Faster R-CNN and YOLO have their own strengths and weaknesses:

- Faster R-CNN is a two-stage detector that first generates region proposals and then classifies and localizes them. In contrast, YOLO is a single-stage detector that predicts bounding boxes and class probabilities simultaneously.

- They use different kinds of backbone networks. Faster R-CNN uses heavyweight networks for feature extraction, contributing to its high accuracy but resulting in slower speed. On the other hand, YOLO employs a lightweight network, enabling faster inference at the cost of slightly lower accuracy.

- When it comes to object detection, Faster R-CNN excels in effectively detecting both small and large objects. On the other hand, YOLO is more efficient at detecting large objects and additionally, it makes fewer errors in background detection.

- In terms of resource requirements, Faster R-CNN demands powerful computational resources, making it more expensive. Conversely, YOLO is less resource-intensive and therefore less expensive.

Given these characteristics, Faster R-CNN is ideally suited for applications that require high accuracy in object detection. Conversely, YOLO, due to its ability to detect the background more accurately, may produce fewer errors in detecting bounding boxes in terms of position. Therefore, in scenarios where the correct positioning of objects is crucial, YOLO might be the better choice. By understanding and leveraging the unique strengths of both models, one can design a more comprehensive object detection system that yields more accurate bounding boxes [10].

## III. FUSION OF FASTER R-CNN AND YOLO

In order to leverage the strengths and mitigate the weaknesses of Faster R-CNN and YOLO for more accurate bounding box predictions, we propose a fusion method. This method utilizes the accurate box predictions of Faster R-CNN and the superior background detection capabilities of YOLO.

The process begins with both Faster R-CNN and YOLO independently predicting bounding boxes for a given image. The fusion method then analyzes these predictions and selects the most appropriate ones (Fig. 3). The selection rules of this method are as follows:

*1) When both models make predictions:* If both YOLO and Faster R-CNN generate predictions for the same image, the bounding boxes from Faster R-CNN are retained only if they encompass the center of a YOLO bounding box. If this condition is not met, the corresponding Faster R-CNN bounding box is discarded.

*2) When only one model makes predictions:* If Faster R-CNN does not predict any bounding boxes for a given image, the bounding boxes predicted by YOLO are used. Conversely, if YOLO does not predict any bounding boxes, all bounding boxes predicted by Faster R-CNN are used.

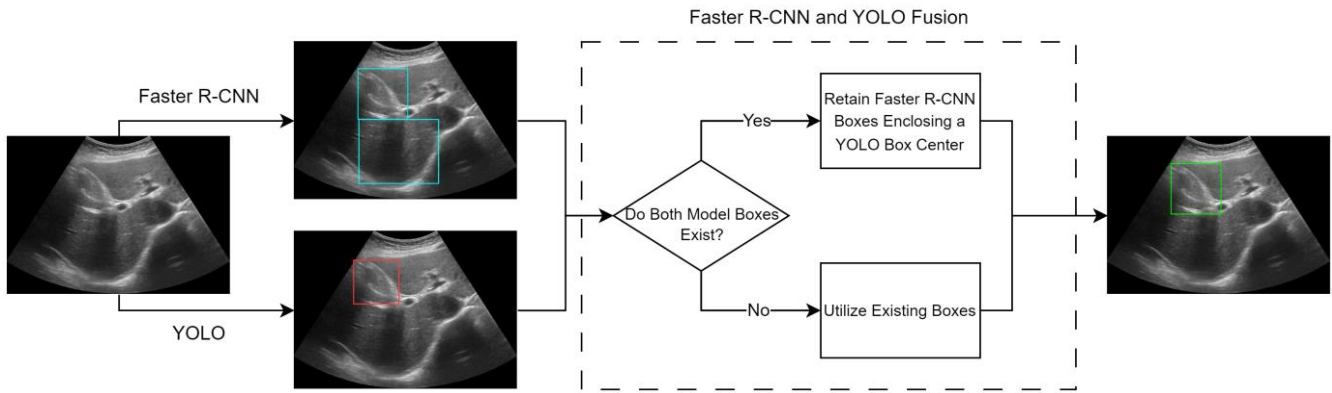

Fig. 3. Illustration of the procedure for the fusion method.

By employing this combination method, the accurate bounding boxes from Faster R-CNN that are correctly positioned will be retained, while those predicting the background will be eliminated. In instances where one of the models fails to predict any boxes, the bounding boxes predicted by the other model will compensate for this shortcoming. Consequently, this approach will yield more precise bounding boxes.

## IV. EXPERIMENTS AND RESULTS

### A. Dataset

The dataset used in this study is the Gallbladder Cancer Ultrasound (GBCU)[1] dataset. Released by Basu et al. [7] in 2022, this dataset is the first of its kind to be made publicly available for the detection of gallbladder cancer from ultrasound images.

The GBCU dataset includes 1255 ultrasound images from 218 patients. These images are categorized as malignant (265 images), benign (558 images), and normal (432 images). The dataset provides a split into a training set with 1133 images and a test set with 122 images. Bounding boxes, which indicate the gallbladder as the region of interest, are also included. This feature makes the dataset suitable for both object detection and classification tasks.

### B. Methodology

In this study, we used the Faster R-CNN and YOLOv8 models to identify the ROIs. The GBCU dataset includes bounding boxes that were obtained by using the Faster R-CNN model. For identifying the ROIs using YOLOv8, we utilized the YOLOv8n model, which is initially pre-trained on the COCO dataset. We further trained this model on the GBCU training set for 100 epochs. A batch size of 72 and SGD as the optimizer, along with the remaining default parameters of YOLOv8 were used. Following the training, we used the model to predict the bounding boxes in the test set. We utilized Basu et al.'s method [7] to assess object detection metrics: Predictions were deemed as true positives if their centers were located within the ground truth box, otherwise they were classified as false positives. The absence of predictions was counted as false negatives. Notably, our evaluation encompassed all predicted boxes, not only the optimal ones.

In the next step, the predicted bounding boxes were used for image classification. For this purpose, we employed the GBCNet classifier, introduced by Basu et al. [7], for GBC detection from ultrasound images. This classifier uses Multi-Scale Second-Order Pooling (MS-SoP) layers. To boost performance, Basu et al. employed a visual acuity training curriculum. The code for the classifier and pre-trained models, both with and without the curriculum training, are available at https://github.com/sbasu276/GBCNet.

This classification model will classify each bounding box in an image. If all of them are classified as normal, the image will be considered normal. If at least one bounding box is classified as malignant, the whole image will be reported as malignant. Otherwise, the image will be labeled as benign. That's why the presence of boxes far from the gallbladder can result in classification errors. For instance, if an image contains a normal gallbladder and one box classifies wrongly as benign or malignant, the whole image will be misclassified to benign or malignant. The same principle applies to an image of a benign gallbladder. If one of the boxes is wrongly predicted as malignant, the entire image is classified as malignant.

In the source code, there are two approaches for training the classifier: using either the ground truth bounding boxes or the predicted bounding boxes from the object detection model. To achieve a more robust classifier model and to eliminate errors from the bounding boxes of the object detection model, using the ground truth bounding boxes during the training phase seems to be a better choice. Therefore, we employed the pre-trained 'GBCNet with Curriculum' model to classify the images. We didn't retrain the model with new predicted bounding boxes each time, as the ground truth of the training dataset remains constant.

For the inference step, there are also two approaches in the source code to handle images with no predicted bounding boxes: using the ground truth bounding box or passing the whole image to the classifier. In this study, we utilized the second approach of handling images with no bounding boxes and passed the whole image to the classifier, as for unseen data, the ground truth bounding boxes might not be provided.

### C. Results and Discussion

Object detection results are summarized in TABLE I. . Based on the results, the following observations can be made:

- The mean Intersection over Union (mIoU) for Faster R-CNN bounding boxes is higher than that for YOLO

---

[1] https://gbc-iitd.github.io/data/gbcu

bounding boxes, indicating that the Faster R-CNN boxes are more accurate and closer to the ground truth.

- Faster R-CNN predicted a greater number of boxes than YOLO, especially in terms of false positives. This indicates that Faster R-CNN generated more bounding boxes that were not correctly positioned.
- YOLO obtained a higher number of true positive boxes than Faster R-CNN, indicating that YOLO positioned boxes more accurately.

Based on these observations, we can conclude that the boxes generated by Faster R-CNN align more accurately with the ground truth. However, Faster R-CNN also produced additional and incorrect boxes, potentially leading to classification errors. In contrast, YOLO didn't generate these extraneous and incorrect boxes, and tends to determine the positions of boxes more accurately. Therefore, it seems reasonable to employ YOLO for selecting the more accurately localized boxes from those generated by Faster R-CNN, and it is anticipated that the proposed fusion method will enhance the results.

TABLE I. OBJECT DETECTION RESULTS

| Methods | Test Set | | | | | |
|---|---|---|---|---|---|---|
| | mIoU | Precision | Recall | TP | FP | FN |
| Faster R-CNN | 69.07 | 87.59 | 98.36 | 120 | 17 | 2 |
| YOLOv8 | 64.85 | 98.44 | 98.44 | 126 | 2 | 2 |
| Faster R-CNN and YOLOv8 Fusion | 74.35 | 92.19 | 99.16 | 118 | 10 | 1 |

mIoU = mean Intersection over Union, TP = True Positive, FP = False Positive, FN = False Negative

As the results clearly demonstrate, compared to the outcomes from Faster R-CNN, the application of the fusion method has led to enhancements in mIoU, Precision, and Recall. This approach has also resulted in a reduction in the number of False Positives and False Negatives. As expected, this indicates that the fusion method produces more precise bounding boxes.

The classification results are summarized in TABLE II. Our findings indicate that the fusion method, which integrates the strengths of both Faster R-CNN and YOLO and modifies the Faster R-CNN bounding boxes, leads to superior classification outcomes. In comparison to the classification based on the unmodified Faster R-CNN bounding boxes, this approach corrected the classifications of three images, which are showcased in Fig. 4.

TABLE II. CLASSIFICATION RESULTS

| Methods Provided Bounding Boxes | Test Set | | | |
|---|---|---|---|---|
| | Acc | 2-Acc | Sens | Spec |
| Faster R-CNN and Ground Truth[a] | 90.98 | 95.90 | 97.62 | 95.00 |
| Faster R-CNN | 90.16 | 95.08 | 95.24 | 95.00 |
| YOLOv8 | 82.79 | 86.89 | 73.81 | 93.75 |
| Faster R-CNN and YOLOv8 Fusion | 92.62 | 96.72 | 97.62 | 96.25 |

Acc = Accuracy, 2-Acc = the binary classification accuracy with malignant and non-malignant classes, as reported in Basu et al.'s paper [7], Sens = Sensitivity, Spec = Specificity

[a] For the images without predicted bounding boxes, ground truth was used

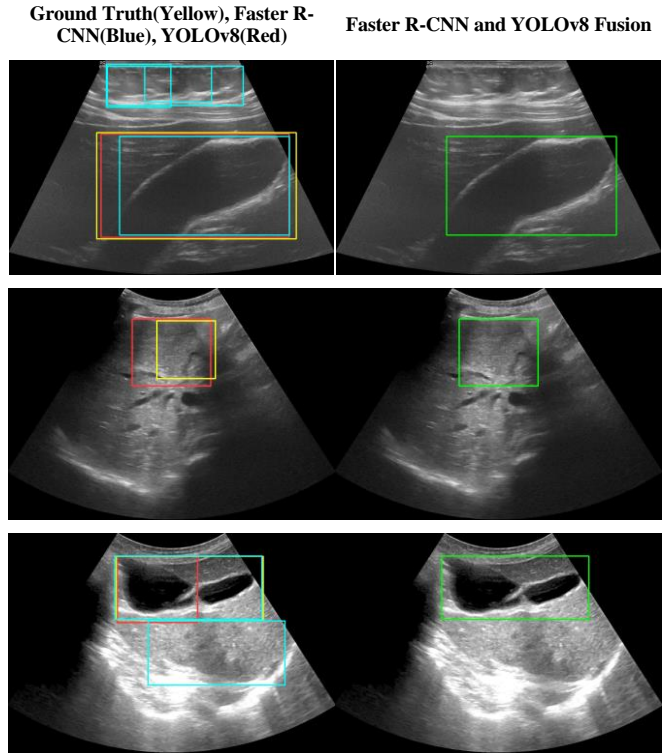

Ground Truth(Yellow), Faster R-CNN(Blue), YOLOv8(Red) | Faster R-CNN and YOLOv8 Fusion

Fig. 4. The fusion method corrected images that were initially misclassified using only Faster R-CNN. This improvement likely resulted from the elimination of incorrectly localized Faster R-CNN bounding boxes and the use of YOLO's bounding boxes in their absence.

D. Error Analysis

Despite the improvements in results using the proposed fusion method, some cases are still classified incorrectly. Fig. 5 displays these misclassified cases.

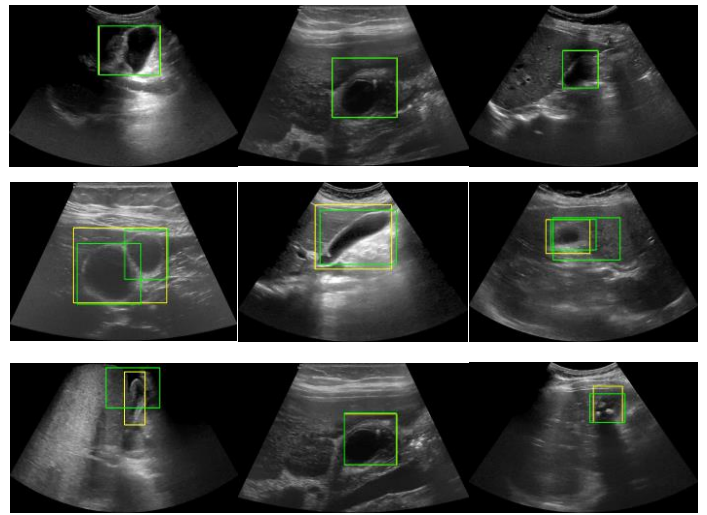

Fig. 5. Misclassified cases, indicated by the fusion method boxes (green) and ground truth boxes (yellow).

When fusion boxes diverge from the ground truth (in almost 56% of cases), classification errors may stem from the object detector. These can be mitigated by refining the detection method. Conversely, when fusion boxes align with the ground truth (in almost 44% of cases) but are misclassified, errors may be due to the classifier or incorrect ground truth annotations. These issues can be addressed by enhancing the classifier and using a robust object detector that may potentially surpass human annotations. Taking these considerations into account, it seems that enhancing the object detector may be more beneficial for future studies.

## V. Discussion

Faster R-CNN and YOLO are two standout techniques in the field of object detection. Each has its own strengths, making them suitable for specific situations. By leveraging the strengths of both models and combining them, a more robust and accurate model can be obtained.

The combination of Faster R-CNN and YOLO has shown improvements in various applications. For instance, Drid et al. [11] enhanced the detection of overlapping and small objects in the PASCAL VOC dataset by retaining YOLO boxes and utilizing Faster R-CNN boxes for objects that were not predicted by YOLO, with this determination made based on IoU. Fan et al. [12] used the Kalman filter to combine Faster R-CNN and YOLOv2 for vehicle detection from video data. Mittal et al. [13] achieved improved results by proposing a model that combines Faster R-CNN and YOLOv5, using majority voting classifier to detect vehicles and estimate traffic density.

In the context of medical imaging, the ability to efficiently and accurately detect regions of interest is of paramount importance. This precision directly enhances the performance of diagnosis and treatment procedures. Faster R-CNN is renowned for its high accuracy, which is vital in medical image object detection to prevent diagnostic errors and mistreatment. Conversely, YOLO excels at detecting the background and tends to make fewer errors in predicting the correct position of objects. This accuracy is also important in medical image detection to avoid confusion. By leveraging the strengths of both models, we proposed a fusion method in this paper that could achieve better results in detecting gallbladder in ultrasound images.

Our findings confirmed that Faster R-CNN was able to detect bounding boxes that were closely aligned with the ground truth. However, it also made errors in detecting the actual gallbladder and incorrectly identified the background in some bounding boxes. While YOLO was less accurate in detecting the boundaries, it was more precise in pinpointing the position of the gallbladder. Our fusion method utilized the YOLO boxes to determine which Faster R-CNN boxes were correctly positioned and eliminated those that were not. This approach resulted in more accurate bounding boxes in terms of both boundaries and position, leading to an improvement in the detection of gallbladder cancer.

This study does have certain limitations. The proposed fusion method is heavily reliant on the individual performance of Faster R-CNN and YOLO. Therefore, it requires fine-tuning and the identification of the best parameters for desired dataset. Additionally, it has only been examined on one dataset and was specifically designed for gallbladder cancer detection. To determine its applicability in a wider range of contexts, further exploration and examinations are necessary.

Looking ahead, future work could focus on developing a unified object detection model that integrates the mentioned strengths of both Faster R-CNN and YOLO. This approach would require fine-tuning of the combined model as a whole, rather than each model separately, potentially leading to a more practical solution. Such a model could prove particularly useful in scenarios where accurate boundary detection with minimal background error is paramount.

## VI. Conclusion

This study introduced a fusion method that integrates the bounding boxes of Faster R-CNN and YOLOv8 for gallbladder detection in ultrasound images, aiming to improve gallbladder cancer diagnosis. Faster R-CNN was able to predict highly accurate bounding boxes, but it also produced multiple bounding boxes that incorrectly identified the background. Conversely, YOLO accurately predicted the position of bounding boxes. By using YOLO boxes to identify and eliminate incorrectly positioned Faster R-CNN boxes, we achieved more accurate bounding boxes. This approach improved the classification results, indicating that our fusion method may offer a promising direction for future research in medical imaging applications.